\documentclass[letterpaper]{article} 
\usepackage{aaai2027}  
\usepackage[hyphens]{url}  
\usepackage{graphicx} 
\usepackage{natbib}  
\usepackage{caption} 
\usepackage[hyphens]{url} 
\usepackage{graphicx} 
\usepackage{natbib} 
\usepackage{caption} 
\usepackage{amsmath}
\usepackage{amssymb}
\usepackage{booktabs}
\usepackage{booktabs}  
\usepackage{tabularx}  
\usepackage{array}     
\usepackage{ragged2e}  
\usepackage{booktabs}
\usepackage{multirow}
\usepackage{amssymb}
\usepackage{enumitem}
\usepackage{makecell}

\usepackage{algorithm}
\usepackage{algorithmic}
\usepackage{booktabs,tabularx,array,ragged2e}
\usepackage{newfloat}
\usepackage{listings}
\DeclareCaptionStyle{ruled}{labelfont=normalfont,labelsep=colon,strut=off} 
\floatstyle{ruled}
\newfloat{listing}{tb}{lst}{}
\floatname{listing}{Listing}

\usepackage{booktabs}

\usepackage{textcomp}

\usepackage{url}
\usepackage{verbatim}
\usepackage{graphicx}
\usepackage{mdframed}
\usepackage{cite}

\usepackage{booktabs}

\usepackage{colortbl}
\usepackage{xcolor}
\usepackage{soul}

\definecolor{Seashell}{RGB}{0, 0, 0} 
\definecolor{Firebrick4}{RGB}{255, 255, 0}

\soulregister{\cite}7
\soulregister{\ref}7
\soulregister{\subref}7

\title{Listen, See and Track: Spatio-Temporal Audio-Visual Sound Event Reasoning\\ for Omni-Modal Language Models}
\author{
   Zhi Zeng\textsuperscript{\rm 1}\equalcontrib,
    Cheng Zhang\textsuperscript{\rm 1}\equalcontrib,
    Zesheng Yang\textsuperscript{\rm 1}\equalcontrib,
    Rendong Pi\textsuperscript{\rm 2}\equalcontrib,
    Jiaying Wu\textsuperscript{\rm 3},
    Di Zhang\textsuperscript{\rm 1},
    Zihan Ma\textsuperscript{\rm 1},
    Guodong Li\textsuperscript{\rm 4},
    Zhou Yang\textsuperscript{\rm 1},
    Yu Xiang\textsuperscript{\rm 2},
    Yifei Zheng\textsuperscript{\rm 5},
    Minnan Luo\textsuperscript{\rm 1}
}
\affiliations{
    \textsuperscript{\rm 1}Xi'an Jiaotong University\\
    \textsuperscript{\rm 2}The Hong Kong Polytechnic University\\
    \textsuperscript{\rm 3}National University of Singapore\\
    \textsuperscript{\rm 4}Central China Normal University\\
    \textsuperscript{\rm 5}University of California San Diego
    
}

\begin{document}
\nocopyright
\maketitle

\begin{abstract}

Understanding dynamic sound sources requires jointly determining what produces a sound, where the source is located, and how it moves over time. Yet existing audio-language models often represent clips as global acoustic events, while vision-language models lack the spatial audio cues needed to localize and track individual sources. To evaluate this missing capability, we introduce \textbf{ST-OmniQA}, a spatio-temporal audio-visual question-answering benchmark built from panoramic videos paired with synchronized first-order Ambisonics (FOA) audio of moving sound sources. It contains 40K videos and 400K question-answer pairs organized into four capability levels covering sound-event recognition, direction of arrival, source distance, motion trajectories, and temporally grounded audio-visual reasoning. Building on this benchmark, we propose \textbf{ST-Omni-R1}, which integrates FOA-derived semantic and trajectory representations with panoramic visual context and is trained through progressive curriculum learning and reasoning-tree reinforcement learning. ST-Omni-R1 achieves 77.83\% average semantic accuracy across the four levels, compared with 37.28\% for the best evaluated baseline. Results on three public spatial-audio benchmarks further indicate that its learned spatial and motion representations transfer beyond ST-OmniQA.
\end{abstract}

\section{Introduction}

Dynamic sound-source understanding is fundamentally a \textbf{tracking and binding problem}: a model must determine what emits a sound, where it is, how it moves, and which visible entity it corresponds to over time. Audio-language models have made substantial progress in audio understanding and reasoning~\citep{Qwen2-audio}, but they typically represent clips through global event semantics and overlook the time-varying geometry of individual sources. Omni-modal language models integrate audio, vision, and language~\citep{gemini,Qwen2.5-omni}, yet their reliance on monaural audio leaves them without the spatial cues required for source localization and trajectory tracking.

Humans resolve this problem by matching auditory and visual evidence across space and time~\citep{stein2008multisensory}. Consider a person walking toward a room where another is seated. Before the walker becomes visible, changes in the direction, intensity, and reverberation of the footsteps reveal the approaching source and its motion~\citep{seifritz2002neural}; once the walker enters view, aligned visual evidence binds the acoustic trajectory to the moving person and distinguishes it from the stationary occupant~\citep{alais2004ventriloquist,parise2012correlation}. The event label \emph{footsteps} and isolated visual frames cannot establish this correspondence, which requires joint reasoning over source semantics, location, motion, and visual identity~\citep{senocak2018learning,owens2018audio,chen2020soundspaces}.

Existing methods address only parts of this problem. Multichannel spatial features have been incorporated into LLMs~\citep{tang2024spatial}, while BAT, OWL, SPUR, and Spatial-Omni support reasoning about source identity, direction, distance, and spatial relations~\citep{bat,owl,spur,spatialomni}; however, they primarily characterize static scenes or clip-level attributes. SELD models recover event activity and time-varying directions, but their fixed vocabularies and structured outputs limit open-ended semantic and relational reasoning~\citep{pseldnets}. Recent audio-only approaches, including Spatial Audio Motion Understanding and Reasoning and concurrent ST-AudioLM, extend spatial modeling to moving-source trajectories~\citep{samur,ST-audio}. ST-AudioLM is especially related to our audio branch through its use of dynamic FOA representations and dense trajectory supervision. Our setting further requires acoustic trajectories to be bound to visible instances and composed with evidence about landmarks, occlusion, and other moving sources. This exposes a \textbf{perception gap} in recovering time-varying source geometry, a \textbf{binding gap} in associating acoustic trajectories with visible entities, and a \textbf{reasoning gap} in composing spatial, temporal, and cross-modal evidence.

To bridge these gaps, we introduce ST-OmniQA, a spatio-temporal audio-visual question-answering benchmark constructed from panoramic videos paired with synchronized first-order Ambisonics (FOA) audio of moving sound sources. ST-OmniQA contains 40K videos and 400K question--answer pairs designed with modality-necessity constraints and executable reasoning graphs. Its four capability levels cover single- and multi-source acoustic perception, spatial and trajectory relations, visual-instance binding, landmark grounding, occlusion understanding, and temporal tracking. We further develop ST-Omni-R1, which integrates semantic and temporally ordered FOA representations from an STA-encoder with panoramic visual context. Progressive curriculum learning introduces the required perceptual and relational capabilities, followed by reasoning-tree reinforcement learning that promotes consistency across intermediate reasoning steps and final answers. Task-level comparisons and stage-wise ablations evaluate its spatio-temporal reasoning capabilities, while experiments on three public spatial-audio benchmarks assess transfer beyond ST-OmniQA.

The main contributions of this paper are:
\begin{itemize}
    \item We introduce \textbf{ST-OmniQA}, a multi-level benchmark containing 40K panoramic videos and 400K question--answer pairs. Modality-necessity constraints and executable reasoning graphs enable controlled evaluation of source perception, multi-source localization, trajectory relations, and compositional audio-visual reasoning.

    \item We develop an STA-encoder for time-varying FOA representations and \textbf{ST-Omni-R1} for integrating sound semantics and temporally ordered trajectories with panoramic visual context. Progressive curriculum learning and reasoning-tree reinforcement learning support increasingly compositional spatio-temporal reasoning.

    \item We conduct task-level comparisons, stage-wise ablations, and transfer evaluations on three public spatial-audio benchmarks, assessing performance on ST-OmniQA and the transfer of learned spatial and motion representations beyond the benchmark.
\end{itemize}

\section{Related Work}

\noindent\textbf{Sound Event Localization and Detection.}
Spatial audio understanding has progressed from sound source localization
(SSL) toward joint localization and event recognition. SSL estimates a
source's direction of arrival (DoA) or three-dimensional position using
classification over discretized directions or regression in angular and
coordinate spaces~\citep{yang2022srp,schymura2021pilot,Zhong2022spherical}.
Range-aware methods further combine DoA with source-to-array distance to
recover full source positions~\citep{diaz2020robust}. Sound event localization
and detection (SELD) extends SSL by jointly predicting event activity and
time-varying source locations. Early systems used convolutional recurrent
networks for joint event recognition and DoA estimation, while SALSA
integrated spectral content and spatial cues within a unified representation
~\citep{adavanne2018sound,Nguyen2022salsa}. PSELDNets further improve
transferability through large-scale synthetic pre-training~\citep{pseldnets}.
Nevertheless, SELD systems generally produce predefined event labels and
frame-level locations, limiting open-ended reasoning about source distance,
long-term trajectories, and cross-modal relations.

\noindent\textbf{Audio-Visual Sound Source Localization and Grounding.}
Audio-visual sound source localization grounds audible events in their
corresponding regions or objects within images and videos. Early
self-supervised methods learned audio-visual correspondence and attention maps
to discover sounding objects without location annotations
~\citep{owens2018audio,senocak2018learning}. Subsequent
work improved object-guided localization and extended visual grounding from
dominant sources to multi-source mixtures and egocentric videos
~\citep{mo2022localizing,hu2022mix,huang2023egocentric}. SoundSpaces and
SoundSpaces~2.0 further combine visual observations with geometrically
consistent acoustics in embodied 3D environments, supporting navigation,
source localization, and separation
~\citep{chen2020soundspaces,chen2022soundspaces2}. Despite this progress, most
methods predict visual regions or task-specific actions rather than binding
time-varying spatial-audio trajectories to multiple visible instances. They
also lack open-ended reasoning about landmark relations, occlusion, and source
motion.

\noindent\textbf{Large Audio-Language Models.}
Large audio-language models (LALMs) align acoustic encoders with large
language models and formulate diverse audio tasks as language generation.
Pengi and LTU support classification, captioning, and open-ended reasoning,
while SALMONN, Qwen2-Audio, Kimi-Audio, and Audio Flamingo extend these
capabilities to general audio interaction, dialogue, and few-shot
adaptation~\citep{pengi,ltu,salmonn,Qwen2-audio,Kimi-audio,audioflamingo}.
However, these models typically represent a clip as global acoustic content
and provide limited access to source direction, distance, and spatial
relations. Tang et al.\ introduce multichannel spatial features into an LLM
for localization and spatially informed speech processing, but the resulting
predictions remain oriented toward task-specific spatial
perception~\citep{tang2024spatial}.

Spatial LALMs bridge this gap by coupling spatial audio encoders with language
models. BAT introduces binaural question answering over sound events,
directions, distances, and source relations~\citep{bat}, whereas OWL adds
geometry-aware supervision and multi-step spatial reasoning~\citep{owl}.
First-order Ambisonics (FOA) approaches such as SPUR and Spatial-Omni further
inject spatial representations into LALMs or omni-modal
models~\citep{spur,spatialomni}. These methods nevertheless emphasize static
or clip-level spatial attributes. Recent work begins to model moving sources:
Spatial Audio Motion Understanding and Reasoning conditions an LLM on
frame-level source tracks~\citep{samur}, while concurrent ST-AudioLM learns
time-resolved FOA representations with dense trajectory
supervision~\citep{ST-audio}. Unlike these audio-only approaches, our method
binds FOA-derived trajectories to synchronized panoramic video and reasons
about visible instances, landmarks, occlusion, and multi-source motion.

\section{ST-OmniQA: Spatio-Temporal Audio-Visual QA Benchmark}
\label{sec:st_Omniqa}

\begin{figure*}[t]
    \centering

    \begin{minipage}[t]{0.245\textwidth}
        \centering
        \includegraphics[width=\linewidth]{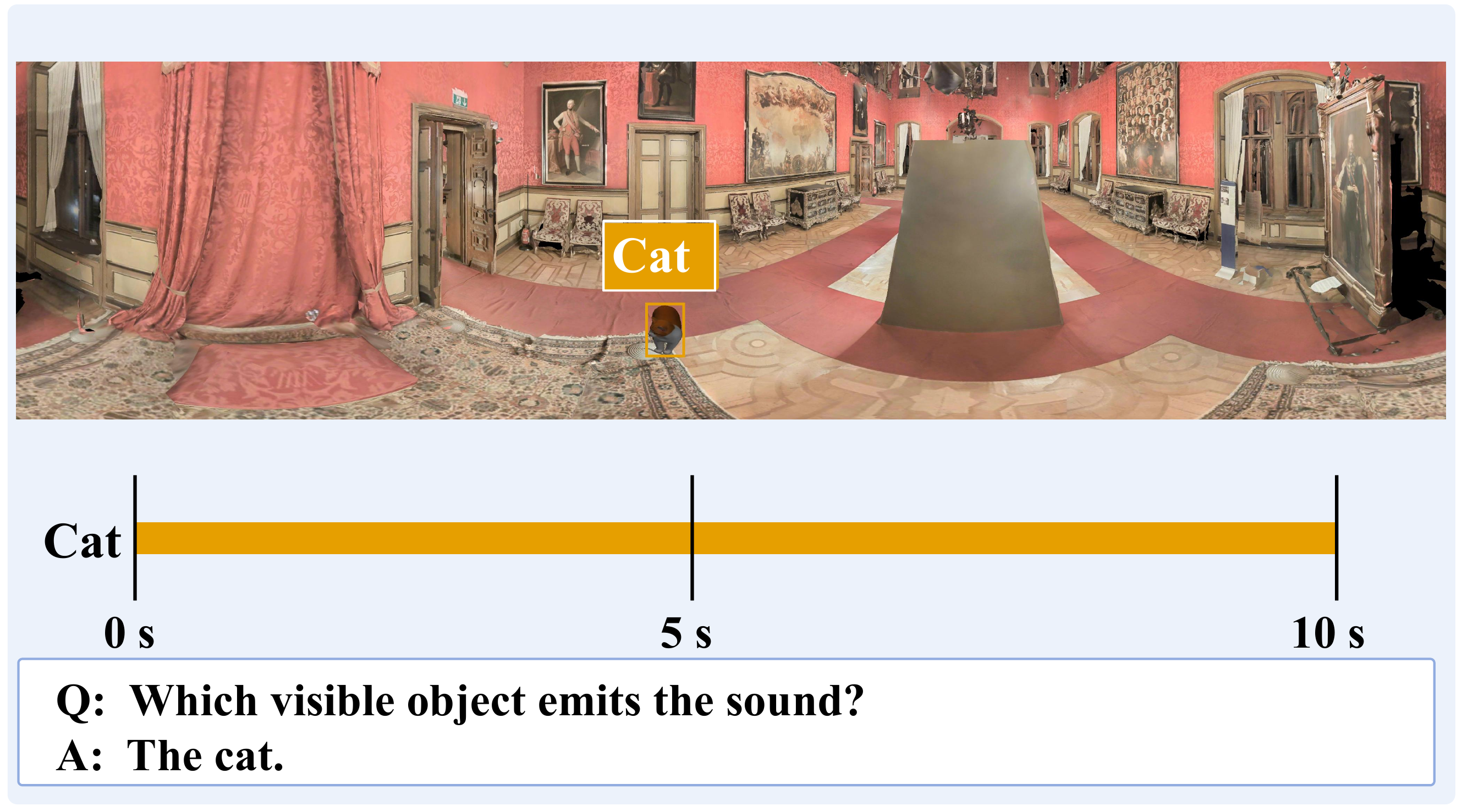}
        \vspace{0.8mm}
        
        \footnotesize\textbf{(a)} Single-source audio-visual grounding
    \end{minipage}\hfill
    \begin{minipage}[t]{0.245\textwidth}
        \centering
        \includegraphics[width=\linewidth]{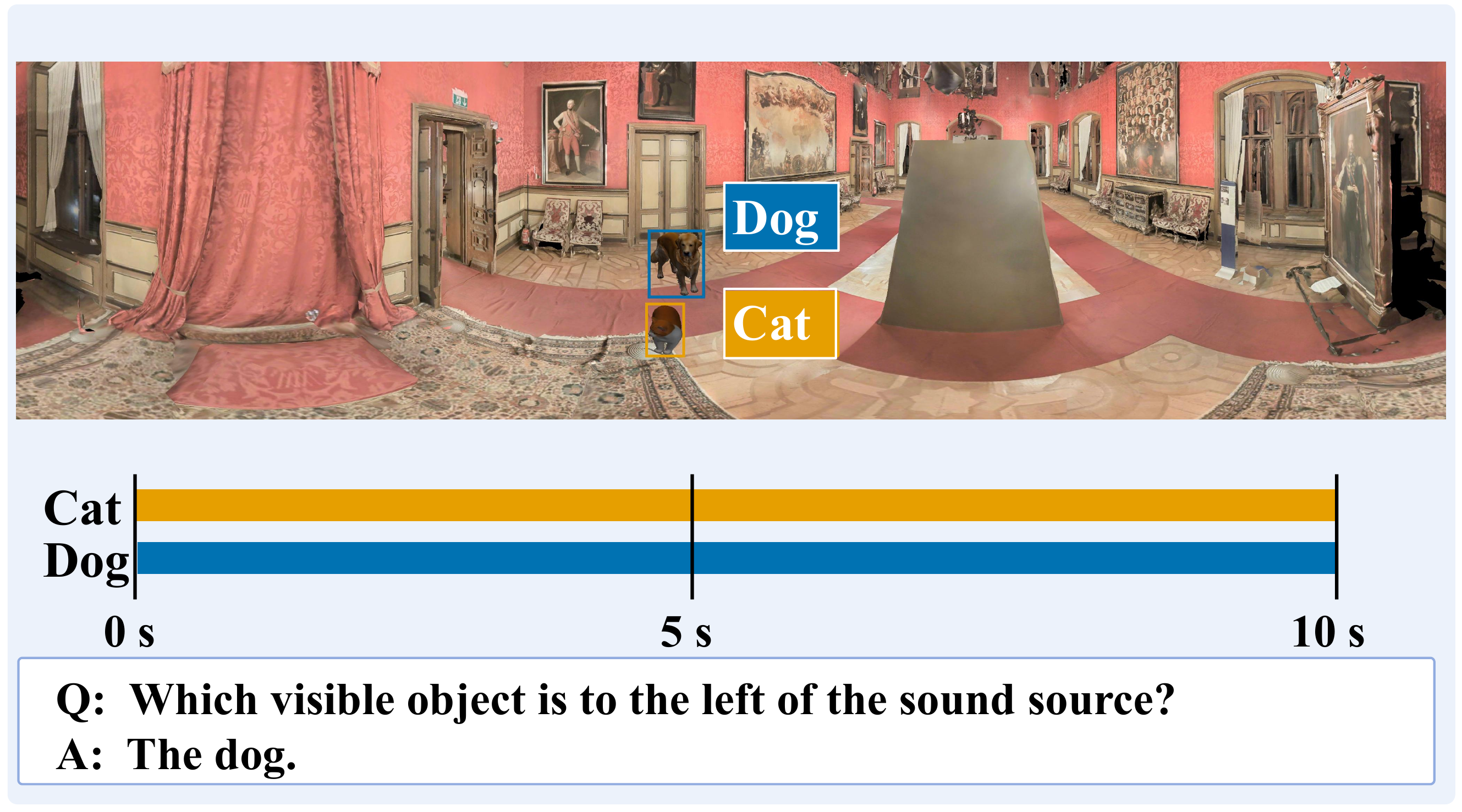}
        \vspace{0.8mm}
        
        \footnotesize\textbf{(b)} Multi-source spatial reasoning
    \end{minipage}\hfill
    \begin{minipage}[t]{0.245\textwidth}
        \centering
        \includegraphics[width=\linewidth]{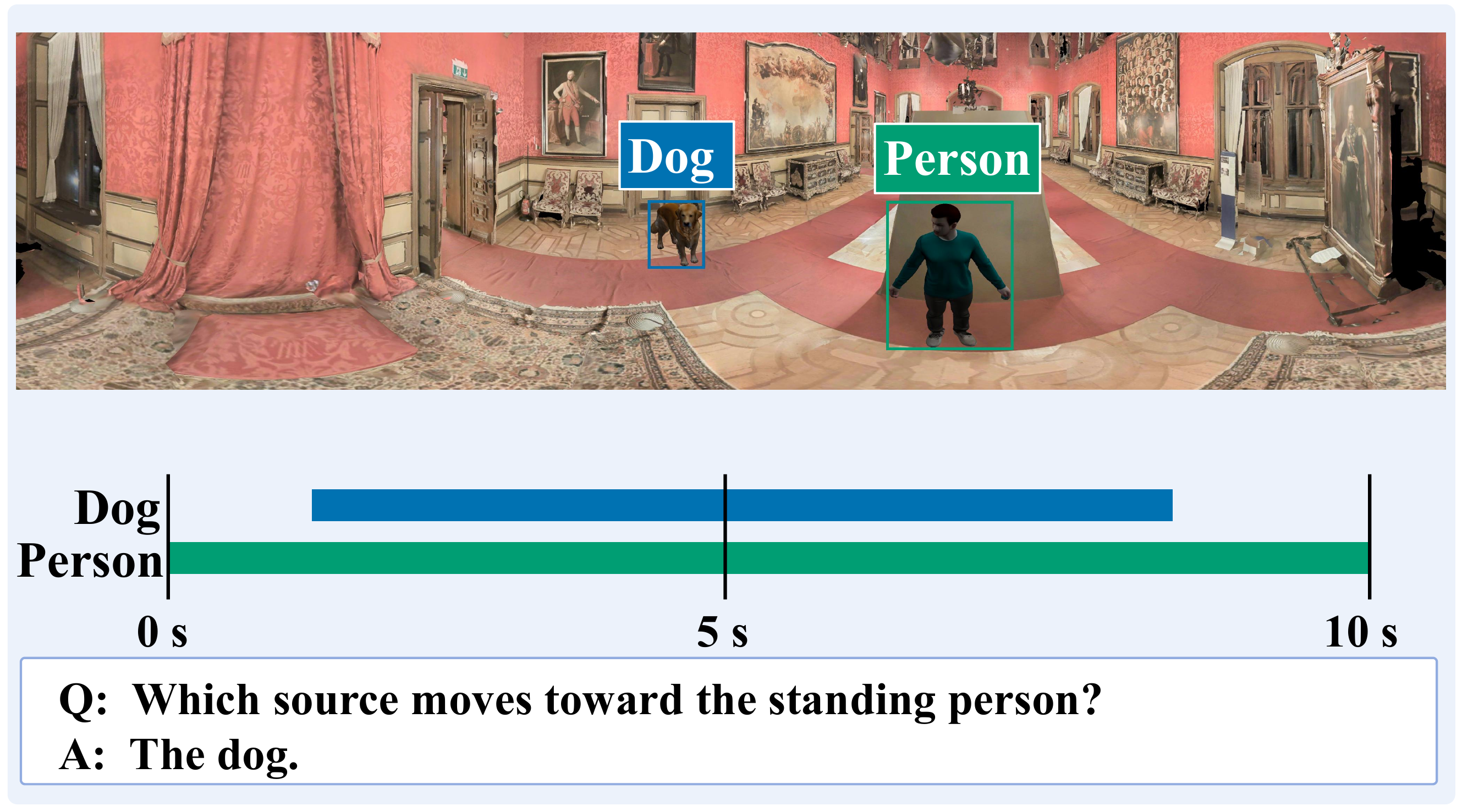}
        \vspace{0.8mm}
        
        \footnotesize\textbf{(c)} Dynamic multi-source event recognition
    \end{minipage}\hfill
    \begin{minipage}[t]{0.245\textwidth}
        \centering
        \includegraphics[width=\linewidth]{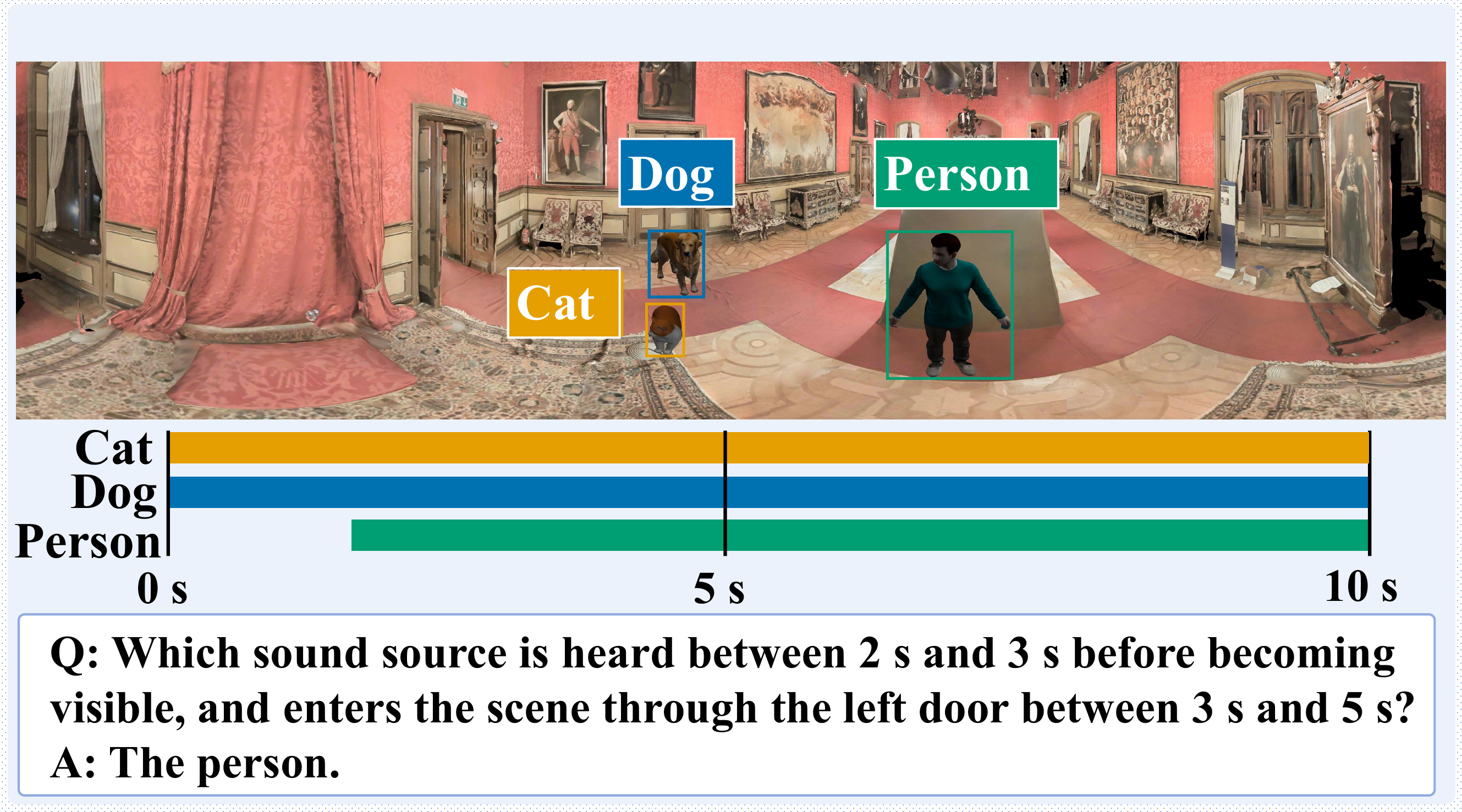}
        \vspace{0.8mm}
        
        \footnotesize\textbf{(d)} Dynamic multi-source trajectory reasoning
    \end{minipage}

    \caption{Representative ST-OmniQA tasks spanning four capability groups:
    (a) single-source audio-visual grounding;
    (b) multi-source spatial reasoning;
    (c) dynamic multi-source event recognition; and
    (d) dynamic multi-source trajectory reasoning.}
    \label{fig:task_examples}
\end{figure*}

\noindent\textbf{Benchmark Formulation.}
ST-OmniQA evaluates complementary capabilities required for scene-grounded
spatio-temporal audio-visual reasoning. Sound-source localization provides
essential spatial cues, but direction or distance alone cannot identify the
visible emitter, distinguish same-class instances, or explain source behavior
during occlusion. The benchmark therefore evaluates acoustic perception,
source-instance binding, and compositional reasoning over source motion and
evolving visual context.

Time-varying acoustic attributes, including source activity, direction,
distance, and event identity, have been investigated in dynamic spatial-audio
QA~\cite{oh2026spatiotemporal}. To support scene-grounded audio-visual
reasoning, ST-OmniQA represents each source using a multimodal state that
jointly captures acoustic, visual, motion, and relational information. For
source $i$ at time $t$, this state is defined as
\begin{equation}
    \mathcal{S}_i(t)=
    \{e_i,a_i(t),\mathbf{g}_i(t),m_i(t),
      v_i(t),o_i,\mathcal{R}_i(t)\},
    \label{eq:benchmark_source_state}
\end{equation}
where $e_i$ denotes the sound-event identity, $a_i(t)$ denotes source
activity, and $\mathbf{g}_i(t)$ contains azimuth, elevation, and
source-listener distance. The variable $m_i(t)$ describes source motion,
whereas $v_i(t)$ and $o_i$ specify visibility and visual-object identity.
Finally, $\mathcal{R}_i(t)$ contains relations to other sources, scene
landmarks, occluding objects, and room structures. Each question is generated
as a deterministic query over either an individual state variable or a
composition of variables across sources, time, and modalities.

\begin{table}[t]
\centering
\setlength{\tabcolsep}{3.2pt}
\renewcommand{\arraystretch}{1.08}
\small
\begin{tabularx}{\columnwidth}{
    >{\centering\arraybackslash}p{8mm}
    >{\raggedright\arraybackslash}p{24mm}
    >{\raggedright\arraybackslash}X}
\Xhline{1.5pt}
\textbf{Level} & \textbf{Capability} & \textbf{Representative tasks} \\
\midrule
A &
Single-source acoustic perception &
Event, activity, DoA, distance, and motion state \\
B &
Multi-source spatial perception &
Target-source selection, disambiguation, and localization \\
C &
Spatio-temporal relational reasoning &
Time-conditioned spatial relations and trajectory comparison \\
D &
Scene-grounded audio-visual reasoning &
Instance binding, landmark grounding, occlusion, and tracking \\
\Xhline{1.5pt}

\end{tabularx}
\caption{Capability progression in ST-OmniQA, spanning source-level acoustic
perception, multi-source localization, spatio-temporal relational reasoning,
and scene-grounded audio-visual reasoning.}
\label{tab:benchmark_levels}
\end{table}

\noindent\textbf{Scene Generation and Annotation.}
ST-OmniQA contains 40K synchronized 10-second panoramic videos with
first-order Ambisonics (FOA) audio, from which we construct 400K
question-answer pairs. Each scene is rendered in a navigable indoor
environment containing one or more sound-emitting 3D objects. The indoor environments are instantiated from Matterport3D scene meshes~\cite{chang2017matterport3d}, while the corresponding room acoustics are simulated with SoundSpaces 2.0~\cite{chen2022soundspaces2}. We consider five
source configurations: single static (\texttt{S}), single dynamic
(\texttt{D}), two static (\texttt{SS}), one static and one dynamic
(\texttt{SD}), and two dynamic (\texttt{DD}).

For moving sources, time-varying room responses are rendered along their
trajectories. This preserves the synchronization among source direction,
distance, reverberation, and visible motion. Panoramic video provides a common
angular reference for the acoustic and visual observations while reducing the
field-of-view ambiguity introduced by conventional perspective cameras.

Each clip is represented by 50 synchronized temporal states. For every source,
we annotate its event class, activity interval, motion state, and time-varying
azimuth, elevation, and distance. Visual annotations include object identity,
screen region, bounding box, visibility state, and the last reliably visible
state. Scene-level annotations additionally record semantic landmarks,
source-to-landmark relations, line-of-sight occluders, room membership, and
room-boundary transitions. These structured annotations support both
listener-centric localization questions and scene-centric questions involving
visible objects, landmarks, trajectories, and occlusion.

\begin{figure*}[t]
    \centering
    \includegraphics[
        width=\textwidth,
        trim=8 5 8 5,
        clip
    ]{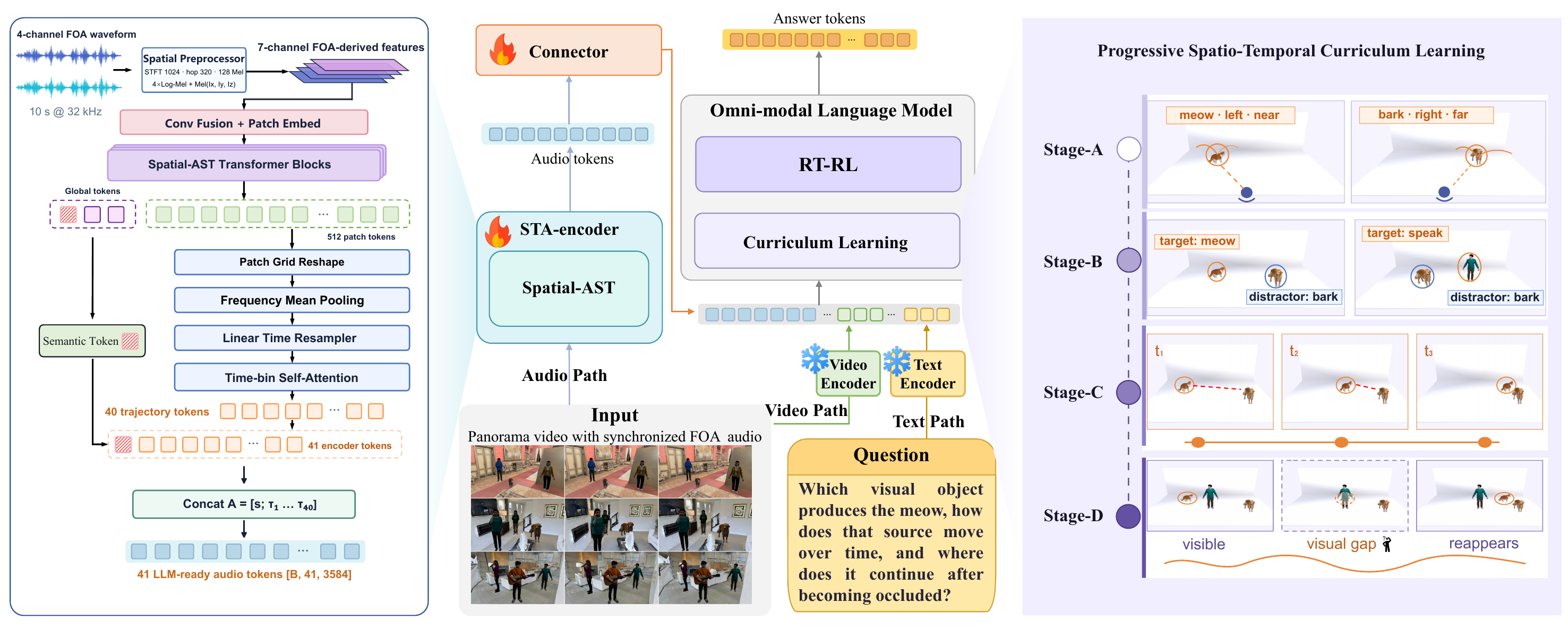}
    \caption{Overview of ST-Omni-R1.}
    \label{fig:overview}
\end{figure*}

\noindent\textbf{Multi-Level Capability Design.}
As summarized in Table~\ref{tab:benchmark_levels}, ST-OmniQA organizes its
questions into four capability levels that differ in source multiplicity,
temporal dependence, relational composition, and cross-modal necessity:
\begin{itemize}[leftmargin=8pt,noitemsep,topsep=0pt]
    \item \textbf{Level A: Single-source acoustic perception.}
    Event identity, activity, direction of arrival (DoA), source-listener
    distance, and motion-state recognition.

    \item \textbf{Level B: Multi-source spatial perception.}
    Target-source selection, instance disambiguation, and localization in
    the presence of competing sources.

    \item \textbf{Level C: Spatio-temporal relational reasoning.}
    Time-conditioned spatial relations, temporal ordering, and trajectory
    comparisons across sources.

    \item \textbf{Level D: Scene-grounded audio-visual reasoning.}
    Source-instance binding, landmark grounding, occlusion understanding,
    visibility transitions, and tracking after visual disappearance.
\end{itemize}
These levels specify complementary evaluation capabilities rather than a
model-independent ranking of empirical difficulty, and are distinct from the
Stage A--D training process. Figure~\ref{fig:task_examples} presents
representative examples for the four levels.





\noindent\textbf{QA Generation and Reasoning Supervision.}
Each question is generated from structured scene states and specifies a target
source, temporal interval, required modalities, and deterministic reasoning
operations. Linguistic templates vary the question form while preserving a
unique answer derived from the annotations.

Levels A and B use answer-only supervision, whereas Levels C and D include
concise traces derived from executable reasoning graphs. Each graph selects
the relevant interval, binds the target source, computes the required
relation, and produces the answer. Privileged simulator metadata is excluded
from these traces.

\noindent\textbf{Modality Necessity and Quality Control.}
For Level D, we retain jointly grounded questions only when neither modality
determines the answer independently. Let $\mathcal{C}_{A}$,
$\mathcal{C}_{V}$, and $\mathcal{C}_{AV}$ denote candidate sets supported by
audio, video, and their joint evidence, respectively:
\begin{equation}
    |\mathcal{C}_{A}|>1,\qquad
    |\mathcal{C}_{V}|>1,\qquad
    |\mathcal{C}_{AV}|=1.
    \label{eq:modality_necessity}
\end{equation}
Thus, the answer becomes unique only after cross-modal binding. Same-class
sources, silent visual distractors, varied activity intervals, and controlled
occlusions reduce unimodal shortcuts. Records are filtered for temporal
alignment, trajectory validity, target uniqueness, stable labels, and exact
reconstructability, and are split by scene-room unit.

\section{ST-Omni-R1: Spatio-Temporal Audio-Visual Reasoning Model}
\label{sec:st_llm_Omni}

\noindent\textbf{Task Formulation and Model Overview.}
Given a panoramic video $\mathbf{V}$, a synchronized first-order Ambisonics
(FOA) waveform $\mathbf{X}^{\mathrm{FOA}}$, and a question $q$,
ST-Omni-R1 models
\begin{equation}
  p_{\boldsymbol{\theta}}
  \!\left(y\mid\mathbf{V},\mathbf{X}^{\mathrm{FOA}},q\right),
  \label{eq:task_formulation}
\end{equation}
where the answer $y$ may require sound-event recognition, direction-of-arrival
(DoA) or source-listener distance estimation, trajectory reasoning, or
temporally aligned visual reasoning.

As shown in Figure~\ref{fig:overview}, the model comprises an STA-encoder, the video encoder of Qwen2.5-VL-7B-Instruct~\citep{qwen2.5vl}, an audio connector, and a multimodal language decoder. The STA-encoder produces one semantic token and $K$ temporally ordered trajectory tokens. The connector maps these tokens into the language-model embedding space for joint decoding with video and text tokens. We first initialize the STA-encoder using static and dynamic FOA supervision. ST-Omni-R1 is then optimized through progressive curriculum learning, followed by reasoning-tree reinforcement learning.

\paragraph{FOA spatial representation.}
FOA contains one omnidirectional component and three directional components. We store the waveform using the AmbiX/ACN convention $(W,Y,Z,X)$ and reorder it as $(W,X,Y,Z)$ when constructing Cartesian spatial features. Let $W_{t,f}$, $X_{t,f}$, $Y_{t,f}$, and $Z_{t,f}$ denote the complex FOA spectra at time $t$ and frequency $f$. We compute the normalized acoustic-intensity
vector
\begin{equation}
  \mathbf{I}_{t,f}
  =
  \frac{
    \operatorname{Re}\!\left(
      W_{t,f}^{*}
      [X_{t,f},Y_{t,f},Z_{t,f}]^{\top}
    \right)
  }{
    |W_{t,f}|^2+|X_{t,f}|^2+|Y_{t,f}|^2+|Z_{t,f}|^2+\epsilon
  }.
  \label{eq:foa_intensity}
\end{equation}

The input feature concatenates channel-wise log-Mel spectrograms and
Mel-projected intensity components:
\begin{equation}
  \mathbf{Z}
  =
  \left[
  \mathbf{M}^{W};
  \mathbf{M}^{X};
  \mathbf{M}^{Y};
  \mathbf{M}^{Z};
  \operatorname{Mel}(I_x);
  \operatorname{Mel}(I_y);
  \operatorname{Mel}(I_z)
  \right].
  \label{eq:foa_feature}
\end{equation}

A channel-fusion layer converts $\mathbf Z$ into a unified time-frequency
representation.

\paragraph{Semantic and trajectory tokens.}
The fused features are divided into non-overlapping time-frequency patches
and processed by an Audio Spectrogram Transformer
(AST)~\citep{gong2021ast}. Learned global tokens encode event semantics,
clip-level DoA, and source-listener distance. The localization tokens provide
auxiliary supervision during STA-encoder initialization, whereas only the
semantic token is retained in the downstream audio interface.
To preserve source motion, we reshape the patch features into a
time-frequency grid
$\mathbf{H}\in\mathbb{R}^{T'\times F'\times d_a}$, average over frequency,
and interpolate the temporal sequence into $K$ ordered bins:
\begin{equation}
  \widetilde{\mathbf{H}}
  =
  \operatorname{Interp}_{K}\!\left(
    \frac{1}{F'}\sum_{f=1}^{F'}\mathbf{H}_{:,f}
  \right)
  \in\mathbb{R}^{K\times d_a}.
  \label{eq:temporal_pooling}
\end{equation}

Temporal self-attention produces trajectory tokens
$\boldsymbol{\tau}_{1:K}$. Auxiliary heads predict source activity
$\hat a_t$, a unit direction vector $\hat{\mathbf d}_t$, and log-distance
$\hat\ell_t$ for each temporal bin. The final audio representation is
\begin{equation}
  \mathbf{A}
  =
  [\mathbf{s};
  \boldsymbol{\tau}_1;
  \ldots;
  \boldsymbol{\tau}_{K}]
  \in\mathbb{R}^{(K+1)\times d_a},
  \label{eq:audio_tokens}
\end{equation}
where $\mathbf{s}$ represents clip-level event semantics and
$\boldsymbol{\tau}_{1:K}$ retain time-varying source states.

\paragraph{Geometric Supervision Alignment.}
In the benchmark state of Eq.~\ref{eq:benchmark_source_state}, let
$\mathbf g_i(t)=(\phi_i(t),\psi_i(t),r_i(t))$ denote azimuth, elevation, and
source-listener distance. Encoder supervision converts this representation
into
\begin{equation}
  \mathbf d_i(t)
  =
  \operatorname{Dir}\!\left(\phi_i(t),\psi_i(t)\right),
  \qquad
  \ell_i(t)=\log r_i(t),
  \label{eq:annotation_mapping}
\end{equation}
where $\operatorname{Dir}(\cdot)$ follows the Cartesian convention used by the
FOA renderer. During single-source encoder initialization, we omit the source
index $i$. The benchmark motion variable $m_i(t)$ is represented by changes
in direction and distance across consecutive temporal bins, while $v_i(t)$,
$o_i$, and $\mathcal R_i(t)$ are supplied through visual tokens and
scene-grounded QA supervision. The 50 benchmark states and the 40 trajectory
tokens serve different purposes: the former define structured QA annotations,
whereas the latter form the time-resolved audio interface to the language
model.

\paragraph{Trajectory-aware initialization.}
The STA-encoder is initialized in two phases before ST-Omni-R1 tuning.
Static-perception pretraining learns event semantics, clip-level DoA, and
source-listener distance. Dynamic adaptation then introduces dense
supervision for activity, direction, and distance over time. When structured
trajectory annotations are used for encoder supervision, they are temporally
aligned with the $K$ encoder bins before computing the objective. For
ground-truth activity $a_t$, direction $\mathbf d_t$, and log-distance
$\ell_t$, the trajectory loss is
\begin{align}
  \mathcal{L}_{\mathrm{traj}}
  =&\;
  \frac{1}{K}\sum_{t=1}^{K}
  \operatorname{BCE}(\hat a_t,a_t)
  \nonumber\\
  &+
  \frac{1}{\sum_t a_t+\epsilon}
  \sum_{t=1}^{K}a_t
  \left(
    \lVert\hat{\mathbf d}_t-\mathbf d_t\rVert_2^2
    +(\hat\ell_t-\ell_t)^2
  \right).
  \label{eq:trajectory_loss}
\end{align}

To reduce the trade-off between event semantics and localization, we retain
the static encoder as a frozen teacher during dynamic adaptation. The 
initialization objective is
\begin{equation}
  \mathcal{L}_{\mathrm{STA}}
  =
  \mathcal{L}_{\mathrm{traj}}
  +
  \lambda_{\mathrm{sem}}
  \left\|
    \hat{\mathbf z}_{\mathrm{sem}}
    -
    \mathbf z^{T}_{\mathrm{sem}}
  \right\|_2^2
  +
  \lambda_{\mathrm{evt}}
  \operatorname{BCE}(\hat{\mathbf e},\mathbf e),
  \label{eq:sta_loss}
\end{equation}
where $\mathbf e$ denotes the event target. After initialization, the
auxiliary heads are removed and the STA-encoder provides the semantic and
trajectory tokens in Eq.~\ref{eq:audio_tokens}.

\noindent\textbf{Spatio-Temporal Audio-Visual Token Fusion.}
\label{sec:token_fusion}
The video encoder transforms the panoramic video into visual tokens
$\mathbf V_{\mathrm{tok}}$. For each audio token $\mathbf a\in\mathbf A$, a
trainable connector computes
\begin{equation}
  g(\mathbf a)
  =
  \operatorname{LN}\!\left(
    \mathbf W_2
    \operatorname{GELU}(\mathbf W_1\mathbf a)
  \right).
  \label{eq:audio_connector}
\end{equation}

The projected audio sequence
$\mathbf A_{\mathrm{tok}}=g(\mathbf A)$ is inserted together with visual and
question tokens into a single decoder context:
\begin{equation}
  \mathbf C
  =
  [\mathbf V_{\mathrm{tok}};
   \mathbf A_{\mathrm{tok}};
   \mathbf Q_{\mathrm{tok}}].
  \label{eq:multimodal_context}
\end{equation}

The decoder can therefore attend jointly to panoramic visual context,
sound-event semantics, ordered source trajectories, and the question. During
model tuning, the pretrained perceptual encoders are frozen, while the
connector and language-side modules align acoustic trajectories with visible
objects and scene relations.

\noindent\textbf{Stage I: Progressive Spatio-Temporal Curriculum Learning.}
\label{sec:curriculum_learning}
As illustrated in Figure~\ref{fig:overview}, we organize curriculum learning
into four progressive stages, denoted as Stage-A, Stage-B, Stage-C, and
Stage-D. The curriculum gradually increases the reasoning difficulty from
single-source acoustic perception to multi-source spatio-temporal analysis
and scene-grounded audio-visual reasoning. Each stage extends the preceding
one with more complex sources, temporal dependencies, and cross-modal
relations.

Given the multimodal context $\mathbf{C}$ and a tokenized response
$\mathbf{y}=(y_1,\ldots,y_L)$, we optimize the model using response-only
next-token cross-entropy:
\begin{equation}
  \mathcal{L}_{\mathrm{SFT}}
  =
  -\sum_{t\in\mathcal I_{\mathrm{resp}}}
  \log p_{\boldsymbol\theta}
  (y_t\mid\mathbf C,y_{<t}),
  \label{eq:sft_loss}
\end{equation}
where $\mathcal I_{\mathrm{resp}}$ denotes the assistant-response positions.
Prompt, padding, video-placeholder, and audio-placeholder tokens are excluded
from the loss.
Stage-A develops single-source event and spatial-motion perception, while
Stage-B extends these capabilities to target identification and localization
in multi-source scenes. Stage-C models temporal and cross-source trajectory
relations, whereas Stage-D incorporates visual evidence for instance binding,
landmark grounding, occlusion, and temporal tracking.

\begin{table*}[t]
  \centering
  \scriptsize
  \setlength{\tabcolsep}{3.9pt}
  \renewcommand{\arraystretch}{1.03}
  \begin{tabular*}{\textwidth}
    {@{\extracolsep{\fill}}lccccccccc@{}}
   \Xhline{1.5pt}
    \multirow{2}{*}{\textbf{Model}}
      & \multicolumn{4}{c}{\textbf{Supported Modalities}}
      & \multicolumn{4}{c}{\textbf{Level Accuracy (\%)}}
      & \multirow{2}{*}{\textbf{Avg.}} \\
    \cmidrule(lr){2-5}
    \cmidrule(lr){6-9}
      & \textbf{S} & \textbf{T} & \textbf{A} & \textbf{V}
      & \textbf{A} & \textbf{B} & \textbf{C} & \textbf{D} & \\
    \midrule
    \multicolumn{10}{c}{\textbf{Open-source Models}} \\
    \midrule
    Qwen2.5-Omni-7B
      & - & \checkmark & \checkmark & \checkmark
      & 13.63 & 11.88 & 43.81 & \underline{32.24} & 27.20 \\
    Qwen3-Omni-30B-A3B
      & - & \checkmark & \checkmark & \checkmark
      & 20.64 & 22.08 & \underline{53.85} & 29.65 & 28.76 \\
    Phi-4-Multimodal-5.6B
      & - & \checkmark & \checkmark & \checkmark
      & 10.02 & 10.62 & 46.15 & 30.05 & 25.32 \\
    Kimi-Audio-7B
      & - & \checkmark & \checkmark & -
      & 16.03 & 14.51 & 47.83 & 16.70 & 18.08 \\
    Audio Flamingo 3
      & - & \checkmark & \checkmark & -
      & 11.82 & 7.15 & 13.04 & 7.87 & 8.44 \\
    \midrule
    \multicolumn{10}{c}{\textbf{Closed-source Models}} \\
    \midrule
    Gemini-2.5-Flash
      & - & \checkmark & \checkmark & \checkmark
      & 34.30 & 25.70 & 49.40 & 24.20 & 33.40 \\
    Gemini-2.5-Pro
      & - & \checkmark & \checkmark & \checkmark
      & \underline{42.10} & \underline{29.50} & 49.80 & 27.70 & \underline{37.28} \\
    Gemini-3.1-Pro
      & - & \checkmark & \checkmark & \checkmark
      & 41.30 & 26.30 & 44.60 & 26.50 & 34.68 \\
    GPT-Audio
      & - & \checkmark & \checkmark & -
      & 16.50 & 12.90 & 46.30 & 23.20 & 24.73 \\
    \midrule
    \multicolumn{10}{c}{\textbf{Spatial-audio Language Model}} \\
    \midrule
    BAT
      & \checkmark & - & \checkmark & -
      & 17.20 & 9.80 & 47.50 & 17.30 & 22.95 \\
    \midrule
    \multicolumn{10}{c}{\textbf{ST-Omni-R1}} \\
    \midrule
    Ours (SFT)
      & \checkmark & \checkmark & \checkmark & \checkmark
      & 81.50 & 74.30 & 54.40 & 90.90 & 75.28 \\
    Ours (SFT+RT-RL)
      & \checkmark & \checkmark & \checkmark & \checkmark
      & \textbf{84.70} & \textbf{76.20} & \textbf{55.70}
      & \textbf{94.70} & \textbf{77.83} \\
   \Xhline{1.5pt}
  \end{tabular*}
   \caption{Performance comparison on ST-OmniQA. Scores are semantic accuracy (\%).
  S, T, A, and V denote spatial cues, temporally ordered evidence, audio,
  and visual inputs, respectively. \textbf{Bold:} the best result. \underline{Underline:} the second result.}
  \label{tab:abcd4k-model-comparison}
\end{table*}

\noindent\textbf{Stage II: Reasoning-Tree Reinforcement Learning.}
\label{sec:reasoning_tree_rl}
Stage~II refines the Stage-D model through reasoning-tree reinforcement
learning, with the objective of enforcing consistency in spatio-temporal
audio-visual reasoning. For each question, a task-specific tree is constructed
from the corresponding ST-OmniQA annotations. The root encodes the multimodal
context, intermediate nodes formalize source states, spatial trajectories,
audio-visual bindings, and scene relations, and terminal nodes specify
candidate answers. Each sampled response is interpreted as a root-to-leaf
reasoning path.

\paragraph{Tree-aware reward.}
The reward integrates format compliance, node-level reasoning consistency, and
final-answer accuracy. Specifically, $r^{\mathrm{fmt}}_j$ and
$r^{\mathrm{ans}}_j$ denote the format and answer scores of response $j$,
while $u_{j,n}\in[0,1]$ measures the correctness of reasoning node $n$.
Violations of parent-child dependencies are quantified by
$p^{\mathrm{con}}_j$. The tree score and overall reward are formulated as
\begin{equation}
\begin{aligned}
  \mathcal{T}_j
  &=
  \frac{1}{|\mathcal N_j|}
  \sum_{n\in\mathcal N_j}u_{j,n}
  -\lambda_{\mathrm{con}}p^{\mathrm{con}}_j,\\
  r_j
  &=
  \lambda_{\mathrm{fmt}}r^{\mathrm{fmt}}_j
  +\lambda_{\mathrm{tree}}\mathcal T_j
  +\lambda_{\mathrm{ans}}r^{\mathrm{ans}}_j,\\
  &\lambda_{\mathrm{fmt}}
  +\lambda_{\mathrm{tree}}
  +\lambda_{\mathrm{ans}}=1.
\end{aligned}
\label{eq:tree_reward}
\end{equation}

Fields with annotation-defined targets are evaluated directly against the corresponding ST-OmniQA annotations, whereas free-form responses are scored using a fixed semantic-equivalence evaluator to accommodate valid linguistic variation.

\paragraph{Group-relative policy optimization.}
Policy optimization follows GRPO~\citep{shao2024deepseekmath}. Given $G$
responses sampled from the same multimodal context $\mathbf C$, the normalized
advantage is
$A_j=(r_j-\overline r_{\mathbf C})/(s_{\mathbf C}+\epsilon)$, where
$\overline r_{\mathbf C}=G^{-1}\sum_{k=1}^{G}r_k$ and $s_{\mathbf C}$ denotes
the within-group reward standard deviation. The token-level policy ratio is
$\rho_{j,t}=\pi_{\boldsymbol\theta}(y_{j,t}\mid\mathbf C,y_{j,<t})/
\pi_{\mathrm{old}}(y_{j,t}\mid\mathbf C,y_{j,<t})$. The policy is optimized
using the following clipped objective:
\begin{equation}
\begin{aligned}
  \mathcal{L}_{\mathrm{RT\text{-}GRPO}}
  ={}&
  -\frac{1}{\sum_j |y_j|}
  \sum_{j,t}
  \min\!\Bigl(
    \rho_{j,t}A_j,\\[-0.2em]
  &\quad
    \operatorname{clip}\!\left(
      \rho_{j,t},1-\varepsilon,1+\varepsilon
    \right)A_j
  \Bigr).
\end{aligned}
\label{eq:rt_grpo_loss}
\end{equation}

Consequently, a non-zero relative optimization signal is obtained only from
groups exhibiting within-group reward variation.

\begin{table*}[t]
  \centering
  \small
  \renewcommand{\arraystretch}{1.03}
  \begin{tabular*}{0.98\textwidth}
    {@{\extracolsep{\fill}}lcccccc@{}}
    \Xhline{1.5pt}
    \textbf{Model}
      & \textbf{Az.} & \textbf{El.} & \textbf{Dist.}
      & \textbf{Mot.} & \textbf{Dir.}
      & \textbf{Overall} \\
    \midrule
    \multicolumn{7}{c}
      {\textbf{TAU-NIGENS SELD 2021}} \\
    \midrule
    BAT
      & 0.00 & \textbf{50.00} & --
      & 0.00 & 0.00
      & 12.50 \\
    \textbf{Ours}
      & \textbf{12.00} & 49.50 & --
      & \textbf{67.50} & \textbf{51.00}
      & \textbf{45.00} \\
    \midrule
    \multicolumn{7}{c}
      {\textbf{L3DAS22 Task 2}} \\
    \midrule
    BAT
      & 0.00 & 50.00 & 16.00
      & -- & --
      & 22.00 \\
    \textbf{Ours}
      & \textbf{11.20} & \textbf{56.00} & \textbf{27.60}
      & -- & --
      & \textbf{31.60} \\
    \midrule
    \multicolumn{7}{c}
      {\textbf{STARSS23}} \\
    \midrule
    BAT
      & 0.00 & 49.70 & 8.98
      & 0.00 & 0.00
      & 11.74 \\
    \textbf{Ours}
      & \textbf{21.56} & \textbf{79.04} & \textbf{20.36}
      & \textbf{84.34} & \textbf{52.41}
      & \textbf{51.54} \\
   \Xhline{1.5pt}
  \end{tabular*}
   \caption{Transferability evaluation on three real-world spatial benchmarks.
Az., El., Dist., Mot., and Dir. denote azimuth, elevation, distance,
motion state, and motion direction, respectively. "--" indicates that
the corresponding capability cannot be evaluated on that dataset. Scores are semantic accuracy (\%).}
   \label{tab:real-spatial-audio-transfer}
\end{table*}


\begin{table}[t]
  \centering
  \footnotesize
  \setlength{\tabcolsep}{3.2pt}
  \renewcommand{\arraystretch}{1.15}
  \begin{tabular}{@{}lrrrrr@{}}
    \Xhline{1.5pt}
    \textbf{Setting}
      & \textbf{A} & \textbf{B} & \textbf{C}
      & \textbf{D} & \textbf{Avg.} \\
    \midrule
    SFT Stage-(A)
      & 79.60 & 50.40 & 4.00 & 12.70
      & 36.68 \\
    SFT Stage-(A+B)
      & 81.40 & 69.40 & 7.20 & 11.10
      & 42.28 \\
    SFT Stage-(A+B+C)
      & {83.30} & {74.50}
      & \textbf{57.80} & 23.50
      & 59.78 \\
    SFT Stage-(A+B+C+D)
      & 81.50 & 74.30 & 54.40
      & {90.90} & {75.28} \\
    \midrule
    Video-only Input
      & 78.90 & 70.50 & 51.60 & 85.50
      & 71.63 \\
    Audio-only Input
      & 35.80 & 31.00 & 52.50 & 68.70
      & 47.00 \\
    \midrule
    Full Model
      & \textbf{84.70} & \textbf{76.20}
      & {55.70} & \textbf{94.70}
      & \textbf{77.83} \\
    \Xhline{1.5pt}
  \end{tabular}
  \caption{Ablation of progressive curriculum learning and input
    modalities on ST-OmniQA. Avg. is the macro-average  semantic accuracy (\%).}
  \label{tab:abcd4k-curriculum}
\end{table}

\section{Experiments}
\label{sec:experiments}



\noindent\textbf{Implementation details.}
We initialize the 12-layer STA-encoder from Spatial-AST and adapt its FOA frontend using 10-second clips spanning 75 event classes and 40-bin source trajectories. Audio is sampled at 32~kHz and represented by four log-Mel channels and three acoustic-intensity features. ST-Omni-R1 uses Qwen2.5-VL-7B-Instruct and samples panoramic video at 2~fps. The STA-encoder produces one semantic token and $K=40$ temporally ordered trajectory tokens. During training, each curriculum stage is trained for one epoch using a learning rate of $10^{-5}$, BF16 precision, and an effective batch size of 128. Stage~II samples eight responses per prompt and uses a learning rate of $10^{-6}$. All experiments are conducted on four NVIDIA A800 GPUs.

\noindent\textbf{Evaluation protocol.}
We evaluate all models across the four difficulty levels of ST-OmniQA, which contains 4,000 questions. To prevent data leakage, the training and test splits are disjoint in both scene identity and temporal interval. We further assess cross-dataset generalization on fixed real-world test subsets from TAU-NIGENS~\citep{politis2021taunigens}, L3DAS22
Task~2~\citep{guizzo2022l3das22}, and STARSS23~\citep{shimada2023starss23}. Unless otherwise specified, all models use greedy decoding; general-purpose baselines follow a unified answer format, whereas BAT retains its original free-form prompting protocol. Semantic accuracy is evaluated by DeepSeek-v4-flash \citep{deepseek-v4}. In the ST-OmniQA evaluation, general-purpose public baselines are assessed without task-specific tuning \footnote{Additional details on the datasets, baselines, chain-of-thought
(CoT) data, and case study are provided in the Appendix.}.

\noindent\textbf{Main Results.}
Table~\ref{tab:abcd4k-model-comparison} compares ST-Omni-R1 with
general-purpose audio-visual models, audio-language models, and spatial-audio language model. ST-Omni-R1 achieves the strongest
overall performance while maintaining a balanced capability profile across
all four levels. General-purpose audio-visual models provide broad semantic
understanding but lack explicit spatial acoustic trajectories, whereas
audio-only models do not directly access visible objects or scene context. By jointly modeling spatial, temporal, acoustic, and visual evidence, ST-Omni-R1 performs particularly well on multi-source grounding and scene-grounded audio-visual reasoning. Stage~II refinement further improves the SFT model across all levels, highlighting the importance of multi-source trajectory reasoning.

\noindent\textbf{Ablation Study.}
Table~\ref{tab:abcd4k-curriculum} examines curriculum progression,
modality contributions, and Stage~II refinement. The curriculum gradually
extends the model from single-source acoustic perception to multi-source
spatio-temporal reasoning and scene-grounded audio-visual understanding.
Modality ablations show that visual input provides strong scene and instance
cues, while audio contributes complementary spatial and temporal information.
Their combination yields a more balanced capability profile across tasks.
Stage~II further consolidates these learned capabilities. Overall, the results
support progressive supervision and joint audio-visual modeling for
spatio-temporal reasoning.

\noindent\textbf{Real-world Transferability Analysis.}
Table~\ref{tab:real-spatial-audio-transfer} evaluates spatial and motion
transfer on three real-world datasets. TAU-NIGENS and L3DAS22 provide audio
only, whereas STARSS23 additionally includes synchronized panoramic video.
ST-Omni-R1 consistently outperforms BAT on the two audio-only benchmarks,
indicating that its spatial and motion representations transfer beyond
ST-OmniQA. Its advantage is more pronounced on STARSS23, where visual evidence
supports binding acoustic trajectories to visible objects. Since BAT lacks a
visual branch, this larger margin reflects performance under each model's
supported modalities rather than controlled audio-only superiority. 

\section{Conclusion}
This work advances audio-visual understanding toward source-level spatio-temporal reasoning. We introduce ST-OmniQA, a large-scale benchmark spanning acoustic perception, localization, trajectory analysis, and visually grounded reasoning, together with ST-Omni-R1, which integrates FOA source dynamics with panoramic visual context through progressive curriculum learning and reasoning-tree reinforcement learning. Experiments show that ST-OmniQA challenges leading models, while ST-Omni-R1 improves semantic, spatial, and compositional reasoning. Its performance on ST-OmniQA and three real-world benchmarks highlights the importance of tracking sound sources as persistent multimodal entities whose semantics, geometry, motion, and visibility evolve over time.

\appendix

\bibliography{aaai2027}


\end{document}